\documentclass[11pt]{article}
\usepackage{pdfpages}
\usepackage{multirow}
\usepackage{array}
\usepackage{graphicx}
\usepackage{latexsym}
\usepackage{pstricks}
\usepackage{pst-node}
\usepackage{pst-tree}
\usepackage{url}
\usepackage{times}
\usepackage{helvet}
\usepackage{courier}
\usepackage{algorithmic}
\usepackage[ruled,vlined,linesnumbered]{algorithm2e}
\usepackage{amsthm}
\usepackage{graphicx,amssymb,amsmath}
\usepackage{mathrsfs}
\usepackage{longtable}

\usepackage{geometry}
\title{An Efficient Implementation for WalkSAT}
\author{ Sixue Liu \\
Institute for Interdisciplinary Information Sciences \\
Tsinghua University\\
Beijing, China \\
sixueliu@gmail.com}

\begin{document}

\maketitle

\begin{abstract}
   Stochastic local search (SLS) algorithms have exhibited great effectiveness in finding models of random instances of the Boolean satisfiability problem (SAT). As one of the most widely known and used SLS algorithm, WalkSAT plays a key role in the evolutions of SLS for SAT, and also hold state-of-the-art performance on random instances. This work proposes a novel implementation for WalkSAT which decreases the redundant calculations leading to a dramatically speeding up, thus dominates the latest version of WalkSAT including its advanced variants.
\end{abstract}

\section{Introduction}

This work is devoted to more efficient implementation for SLS algorithm based on focused random walk framework. We propose a new scheme called separated-non-caching to compute the $break$ value, which decreases some unnecessary calculations and improves the efficiency. Combining all these, we design a new SAT solver dubbed WalkSNC (WalkSAT with separated-non-caching).

The experimental results show that WalkSNC significantly outperforms the latest version of WalkSAT including its state-of-the-art variants, especially on large scale benchmarks. SAT Competition has been held for more than 10 years to evaluate state-of-the-art SAT solvers. Our benchmark includes random $k$-SAT instances on the phase transition point from SAT Competition 2013 and 2014, and many larger instances generated by the uniform random $k$-SAT generator.

The rest of this paper is organized as follows. Some necessary notations and definitions are given in the next section, then the separated-non-caching technology is introduced. And experimental evaluations are illustrated after that. Finally, we give conclusions of this work and future directions.

\newtheorem{Definition}{Definition}
\newtheorem{Lemma}{Lemma}
\newtheorem{Remark}{Remark}

\section{Preliminaries}

Given a Conjunctive Normal Form(CNF) formula $F=c_1 \bigwedge...\bigwedge c_m$ on a variables' set $V=\{v_1,v_2,...,v_n\}$, where $c_i$ is a clause and consists of literals: boolean variables or their negations. A $k$-SAT formula is a CNF where each clause contains at most $k$ literals. The $ratio$ of a CNF is defined as the ratio of the the number of clauses and the number of variables. An assignment $\alpha$ is called complete if it matches every variable with TRUE or FALSE. We say a literal is a true literal if it evaluates to TRUE under $\alpha$. The task of the SAT problem is to answer whether there exists a complete assignment such that all clauses are satisfied.

SLS algorithms under focused random walk framework first choose an unsatisfied clause $c$, then choose a flip variable from $c$ according to some rules (Algorithm 1).
These rules are usually based on variables' information like $break$ and $make$.
The $break$ value of $v$ is the number of clauses which will become unsatisfied from satisfied after flipping $v$. While the $make$ value is the number of clauses which will become satisfied from unsatisfied after flipping $v$.
A traditional SLS algorithm called WalkSAT/SKC uses a simple rule to pick variable: if there exists a variable with $break=0$, flip it, otherwise flip a random variable with probability $p$, or a variable with minimal $break$ with probability $1-p$.
Another SLS algorithm called probSAT also uses $break$ value only, but in a completely probabilistic way.
Some recent SLS algorithms also utilize some other information of variables to obtain more complex rules \cite{cai2012configuration}: the $neighborhood$ of a variable $v$ are all the variables that occur in at least one same clause with $v$. The $score$ of a variable is defined as the sum
of weights of clauses (at least 1) which will become satisfied from unsatisfied after flipping that variable. If variable $v$'s neighborhood has been flipped since $v$'s last flip, $v$ is called $configuration\; changed$ variable, and $configuration\; change\; decreasing$(CCD) variables if $score(v)>0$ too.
This notion has a significant influence to state-of-the-art SLS algorithms, and we will illustrate the connection between our algorithm and it.

\begin{algorithm}[t]
\KwIn{CNF-formula $F$, $maxSteps$}
\KwOut{A satisfying assignment $\alpha$ of $F$, or $Unknown$}
\Begin{
    $\alpha$ $\leftarrow$ random generated assignment;\\
    \For {$step$ $\leftarrow$ \upshape{1} \bf{to} $maxSteps$}{
            \textbf{if $\alpha$ $satisfies$ $F$ then return $\alpha$};\\
            $c$ $\leftarrow$ an unsatisfied clause chosen randomly;\\
            $v$ $\leftarrow$ $pickVar(c)$ \\
            $\alpha$ $\leftarrow$ $\alpha$ with $v$ flipped; \\
    }
    \Return $Unknown$
}
\caption{ Focused Random Walk Framework
\label{up}
}
\end{algorithm}

\subsection{Separated-non-caching Technology}

Implementation affects the performances of SLS algorithm very much. The latest version of probSAT uses caching scheme with XOR technology \cite{balintimproving}, while WalkSAT in UBCSAT framework \cite{tompkins2005ubcsat} and the latest version of WalkSATlm \cite{cai2014improving} are under non-caching implementation. In this section, we propose a more efficient implementation called separated-non-caching. The 'separated' term means separated the non-caching process of calculating $break$, to find 0-break variables as soon as possible to reduce unnecessary calculations.

Recall the caching scheme updates every information including the $break$ value of each variables. However, if there exists variable with $break=0$, the other variable's $break$ value is useless. We try to reduce the wasting calculations and only compute what we need.

Under our new implementation, the $flip$ operation only updates the unsatisfied clauses' set and the true literal numbers of every clause, but leave the $break$ calculation to $pickVar$ function.
There are some necessary definitions to compute $break$ value.

\begin{Definition}
        For each clause $c$, $NT(c)$ donates the number of true literals in $c$.
\end{Definition}

$NT(c)=0$ means $c$ is unsatisfied, satisfied clause always has positive $NT$.

\begin{Definition}
        For each variable $v$, $TLC(v)$ donates all the clauses containing the true literal $v$ if $v=TRUE$ under the current assignment or $\bar{v}$ vice versa.
\end{Definition}

If $v$ is TRUE under the current assignment, all clauses contains positive $v$ become $TLC(v)$. Else if $v$ is FALSE, all the clauses contains negative $v$ are $TLC(v)$. All $c$ in $TLC(v)$ with only one true literal will contribute 1 to $break(v)$. Because $v$ is the only one true literal in $c$, flipping $v$ will falsify this literal and make $c$ unsatisfied. We need an additional boolean flag $zero$ to donate whether 0-break variables exist.

\begin{algorithm}[t]
\KwIn{An unsatisfied clause $c$}
\KwOut{A variable $v$ $\in$ $c$}
\Begin{
        Generate a random order of all the variables in $c$; \\
        \ForEach {$v$ $\in$ $c$} {
                Initiate all the clauses in $TLC(v)$ as $unvisited$; \\
                $zero$ $\leftarrow$ $TRUE$; \\
                \ForEach {$ci$ $\in$ $TLC(v)$} {
                        mark $ci$ as $visited$ in $TLC(v)$;\\
                        \If {$NT(ci)=1$} {
                                $zero$ $\leftarrow$ $FALSE$;\\
                                \textbf{break};
                        }
                }
                \If {$zero$=TRUE} {
                        \Return $v$;
                }
        }
        With probability 0.567, \Return a randan chosen variable in $c$;\\
        Initialize the $bestVar$ as the first variable;\\
        \ForEach {$v$ $\in$ $c$} {
                $break(v)$ $\leftarrow$ $1$;\\
                \ForEach {unvisited $ci$ $\in$ $TLC(v)$} {
                        \If {$NT(ci)=1$} {
                                $break(v)\leftarrow break(v)+1$;\\
                                \If {$break(v) \ge break(bestVar)$} {
                                    \textbf{break};
                                }
                        }
                }
                $bestVar \leftarrow v$;\\
        }
        \Return the variable $bestVar$;
}
\caption{ The Implementation of $pickVar$ Function of separated-non-caching
\label{up}
}
\end{algorithm}

In the separated-non-caching implementation outlined in algorithm 2, if there are more than one 0-break variables, return a random one. So in line 2, we first generate a random order to guarantee the first variable with $break=0$ is a random 0-break variable. That's why line 11 can directly return a random 0-break variable. Line 3 to line 12 is to decide whether exists 0-break variable.
The condition in line 8 implies the variable's break value is at least 1, thus $zero$ is marked as False and the algorithm switches to another variable.

If 0-break variable doesn't exist, return the a random variable with probability 0.567, or return the variable with minimal $break$ value with the remaining probability. If the currently break value of the variable reaches or exceeds the best variables with minimal break value, the rest of the unvisted clauses don't have to be numerated, so line 20 is also an efficient pruning.
If every clause is visited, then this implies the break value is smaller, then the $bestVar$ can be updated.

Because we mark the clauses in $TLC$, so at most $|c| \times |TLC|$ clauses are visited. The average size of $TLC$ is $k \times ratio/2$, $k$ donate $k$-SAT. For random 3-SAT with $ratio=4.2$, it's about $3 \times 3 \times 4.2/2=18.9$ clauses to visit. However, due to the existence of 0-break variable, the average visited clauses are much less than 18.9.

\section{Experimental Evaluations}

\begin{table*}
\centering
\begin{tabular}{|c||c|c|c|c|}\hline

\multirow{3}{1.7cm}{\centering Instance\\ Class} &
\multirow{3}{1.9cm}{\centering WalkSATv51 \\ suc \\ par10 } &
\multirow{3}{1.9cm}{\centering WalkSATlm \\ suc \\ par10 }  & \multirow{3}{1.9cm}{\centering probSAT \\ suc \\ par10 }  & \multirow{3}{1.9cm}{\centering WalkSNC \\ suc \\ par10 } \\
&&&&\\&&&&\\ \hline \hline

\multirow{2}{1.9cm}{\centering SC13} & \multirow{2}{1.9cm}{\centering 10.2\% \\ 45340 } & \multirow{2}{1.9cm}{\centering 26.4\% \\ 36980} & \multirow{2}{1.9cm}{\centering 29.1\% \\ 35010  } & \multirow{2}{1.9cm}{\centering \textbf{37.4\%} \\ \textbf{31198} }  \\
&&&&\\ \hline

\multirow{2}{1.9cm}{\centering SC14} & \multirow{2}{1.9cm}{\centering 9.6\% \\ 45120 } & \multirow{2}{1.9cm}{\centering 27.1\% \\ 36603 } & \multirow{2}{1.9cm}{\centering 19.0\% \\ 40994 } & \multirow{2}{1.9cm}{\centering \textbf{35.5\%} \\ \textbf{33010} }  \\
&&&&\\ \hline

\multirow{2}{1.9cm}{\centering V-$10^5$} & \multirow{2}{1.9cm}{\centering 95.3\% \\ 3453 } & \multirow{2}{1.9cm}{\centering 99.0\% \\ 1292 } & \multirow{2}{1.9cm}{\centering 100\% \\ 763 } & \multirow{2}{1.9cm}{\centering \textbf{100\%} \\ \textbf{432} }  \\
&&&&\\ \hline

\multirow{2}{1.9cm}{\centering V-$10^6$} & \multirow{2}{1.9cm}{\centering 89.2\% \\ 6159 } & \multirow{2}{1.9cm}{\centering 98.0\% \\ 2274 } & \multirow{2}{1.9cm}{\centering 99.1\% \\ 1514 } & \multirow{2}{1.9cm}{\centering \textbf{99.8\%} \\ \textbf{499} }  \\
&&&&\\ \hline

\end{tabular}
\caption{Comparison on random 3-SAT}
\end{table*}

We carry out large-scale experiments to evaluate WalkSNC on random $k$-SAT instances at the phase transition point.

\subsection{The Benchmarks}

We adopt 3 random random benchmarks from SAT competition 2013 and 2014 as well as 100 instances we generated randomly. The experiments for $k$-SAT ($k>3$) are not reported here but will be shown in the full version.

\begin{itemize}
\item SC13: 50 different variables instances with $ratio = 4.267$. From the threshold benchmark of the random SAT track of SAT competition 2013\footnote{http://www.satcompetition.org/2013/downloads.shtml/Random Benchmarks}.

\item SC14: 30 different variables instances with $ratio = 4.267$. From the threshold benchmark of the random SAT track of SAT competition 2014\footnote{http://www.satcompetition.org/2014/downloads.shtml/Random Benchmarks}.

\item V-$10^5$: Generated by the SAT Challenge 2012 generator with 50 instances for 100,000 variables, $ratio = 4.2$.

\item V-$10^6$: Generated by the SAT Challenge 2012 generator with 50 instances for 1,000,000 variables, $ratio = 4.2$.

\end{itemize}

Note that the instances from SAT Competition 2013 and 2014 have approximately half unsatisfied fraction.

\subsection{The Competitors}

We compare WalkSNC with the latest version of WalkSAT downloaded from WalkSAT homepage \footnote{https://www.cs.rochester.edu/u/kautz/walksat/}, and a state-of-the-art implementation based on non-caching WalkSATlm, and its variant probSAT which is the championship of SAT competition 2013 random track.

\subsection{Evaluation Methodology}

The cutoff time is set to 5000 seconds as same as in SAT Competition 2013 and 2014, which is enough to test the performance of SAT solvers. Each run terminates finding a satisfying within the cutoff time is a successful run. We run each solver 10 times for each instance from SAT Competition 2013 and 2014 and thus 500 runs for each class. We report ``suc" as the ratio of successful runs and total runs, as well as the ``par10" as the penalized average run time(a unsuccessful run is penalized as $10 \times$ cutoff time). The result in \textbf{bold} indicates the best performance for a class.

All the experiments are carried out on our machine with Intel Core Xeon E5-2650 2.60GHz CPU and 32GB memory under Linux.

\subsection{Experimental Results}

Table 1 shows the comparative results of WalkSNC and their state-of-the-art competitors on the 3-SAT threshold benchmark. The best performance is achieved by WalkSNC, the others performs relatively poor.
Considering the constant speeding up, the average time over all the successful runs of WalkSATv51 is almost 1.5 times of WalkSNC, and WalkSATlm is also 25\% slower than us. The comparison of data we report on par10 is even more distinct.

\section{Conclusions and Future Work}

This work opens up a totally new research direction in improving SLS for SAT: instead of calculating and utilizing extra and precise information of variables to decide which one to be flipped, there are much more things to dig using the simple information. What matters most is balancing the cost of calculating them and the benefits they bring. Additionally, this new technology can be easily adapted to new algorithms based on WalkSAT and probSAT.



\bibliographystyle{plain}
\bibliography{TR}

\end{document}